\documentclass[conference]{IEEEtran}
\usepackage{times}

\usepackage{amsmath}
\usepackage{amsfonts,amssymb}
\usepackage{mathtools}
\usepackage{bm}
\usepackage{siunitx}
\usepackage{amsthm}
\usepackage{verbatim}
\usepackage{breqn}
\usepackage{siunitx}
\usepackage{subfig}
\usepackage{booktabs} 
\usepackage[ruled,vlined]{algorithm2e}

\SetCommentSty{mycommfont}
\usepackage[pagebackref=true,breaklinks=true,letterpaper=true,colorlinks,bookmarks=false]{hyperref}

\usepackage[numbers]{natbib}
\usepackage{multicol}
\usepackage{hyperref}

\usepackage{xcolor}

\newcommand{\shuran}[1]{\textcolor{red}{[Shuran says: #1]}}

\begin{document}
\def\Blue{\color{blue}}
\def\Purple{\color{purple}}

\def\A{{\bf A}}
\def\a{{\bf a}}
\def\B{{\bf B}}
\def\b{{\bf b}}
\def\C{{\bf C}}
\def\c{{\bf c}}
\def\D{{\bf D}}
\def\d{{\bf d}}
\def\E{{\bf E}}
\def\e{{\bf e}}
\def\f{{\bf f}}
\def\F{{\bf F}}
\def\K{{\bf K}}
\def\k{{\bf k}}
\def\L{{\bf L}}
\def\H{{\bf H}}
\def\h{{\bf h}}
\def\G{{\bf G}}
\def\g{{\bf g}}
\def\I{{\bf I}}
\def\R{{\bf R}}
\def\X{{\bf X}}
\def\Y{{\bf Y}}
\def\OO{{\bf O}}
\def\oo{{\bf o}}
\def\P{{\bf P}}
\def\Q{{\bf Q}}
\def\r{{\bf r}}
\def\s{{\bf s}}
\def\S{{\bf S}}
\def\t{{\bf t}}
\def\T{{\bf T}}
\def\x{{\bf x}}
\def\y{{\bf y}}
\def\z{{\bf z}}
\def\Z{{\bf Z}}
\def\M{{\bf M}}
\def\m{{\bf m}}
\def\n{{\bf n}}
\def\U{{\bf U}}
\def\u{{\bf u}}
\def\V{{\bf V}}
\def\v{{\bf v}}
\def\W{{\bf W}}
\def\w{{\bf w}}
\def\0{{\bf 0}}
\def\1{{\bf 1}}
\def\N{{\bf N}}

\def\AM{{\mathcal A}}
\def\EM{{\mathcal E}}
\def\FM{{\mathcal F}}
\def\TM{{\mathcal T}}
\def\UM{{\mathcal U}}
\def\XM{{\mathcal X}}
\def\YM{{\mathcal Y}}
\def\NM{{\mathcal N}}
\def\OM{{\mathcal O}}
\def\IM{{\mathcal I}}
\def\GM{{\mathcal G}}
\def\PM{{\mathcal P}}
\def\LM{{\mathcal L}}
\def\MM{{\mathcal M}}
\def\DM{{\mathcal D}}
\def\SM{{\mathcal S}}
\def\RB{{\mathbb R}}
\def\EB{{\mathbb E}}

\def\tx{\tilde{\bf x}}
\def\ty{\tilde{\bf y}}
\def\tz{\tilde{\bf z}}
\def\hd{\hat{d}}
\def\HD{\hat{\bf D}}
\def\hx{\hat{\bf x}}
\def\hR{\hat{R}}

\def\Ome{\mbox{\boldmath$\omega$\unboldmath}}
\def\bet{\mbox{\boldmath$\beta$\unboldmath}}
\def\et{\mbox{\boldmath$\eta$\unboldmath}}
\def\ep{\mbox{\boldmath$\epsilon$\unboldmath}}
\def\ph{\mbox{\boldmath$\phi$\unboldmath}}
\def\Pii{\mbox{\boldmath$\Pi$\unboldmath}}
\def\pii{\mbox{\boldmath$\pi$\unboldmath}}
\def\Ph{\mbox{\boldmath$\Phi$\unboldmath}}
\def\Ps{\mbox{\boldmath$\Psi$\unboldmath}}
\def\pss{\mbox{\boldmath$\psi$\unboldmath}}
\def\tha{\mbox{\boldmath$\theta$\unboldmath}}
\def\Tha{\mbox{\boldmath$\Theta$\unboldmath}}
\def\muu{\mbox{\boldmath$\mu$\unboldmath}}
\def\Si{\mbox{\boldmath$\Sigma$\unboldmath}}
\def\Gam{\mbox{\boldmath$\Gamma$\unboldmath}}
\def\gamm{\mbox{\boldmath$\gamma$\unboldmath}}
\def\Lam{\mbox{\boldmath$\Lambda$\unboldmath}}
\def\De{\mbox{\boldmath$\Delta$\unboldmath}}
\def\vps{\mbox{\boldmath$\varepsilon$\unboldmath}}
\def\Up{\mbox{\boldmath$\Upsilon$\unboldmath}}
\def\Lap{\mbox{\boldmath$\LM$\unboldmath}}
\newcommand{\ti}[1]{\tilde{#1}}

\def\tr{\mathrm{tr}}
\def\etr{\mathrm{etr}}
\def\etal{{\em et al.\/}\,}
\newcommand{\indep}{{\;\bot\!\!\!\!\!\!\bot\;}}
\def\argmax{\mathop{\rm argmax}}
\def\argmin{\mathop{\rm argmin}}
\def\vec{\text{vec}}
\def\cov{\text{cov}}
\def\dg{\text{diag}}

\newcommand{\tabref}[1]{Table~\ref{#1}}
\newcommand{\lemref}[1]{Lemma~\ref{#1}}
\newcommand{\thmref}[1]{Theorem~\ref{#1}}
\newcommand{\clmref}[1]{Claim~\ref{#1}}
\newcommand{\crlref}[1]{Corollary~\ref{#1}}
\newcommand{\eqnref}[1]{Eqn.~\ref{#1}}

\newtheorem{remark}{Remark}
\newtheorem{theorem}{Theorem}
\newtheorem{lemma}{Lemma}
\newtheorem{definition}{Definition}

\newtheorem{proposition}{Proposition}

\newcommand{\CZ}[1]{{\color{blue}[Chengzhi says: #1]}}
\newcommand{\RL}[1]{{\color{red}[Ruoshi says: #1]}}
\newcommand{\red}[1]{\color{red}[#1]}


\title{\vspace{-0.4cm}PaperBot:\\Learning to Design Real-World Tools Using Paper}

\author{Ruoshi Liu$^1$ \ \ Junbang Liang$^1$ \ \ Sruthi Sudhakar$^1$ \ \ Huy Ha$^{1,2}$ \ \ Cheng Chi$^{1,2}$ \ \  Shuran Song$^{2}$ \ \ Carl Vondrick$^1$ 
\vspace{0.1cm}
\\$^1$
\hspace{.1cm}Columbia University \ \ $^2$\hspace{.1cm}Stanford University
\vspace{0.08cm}\\
{\href{https://paperbot.cs.columbia.edu}{{\url{paperbot.cs.columbia.edu}}}}
\vspace{-0.2cm}
\\}


%

\twocolumn[{%
            \renewcommand\twocolumn[1][]{#1}%
            \maketitle
            \begin{center}
                \vspace{-0.18cm}
                \includegraphics[width=\linewidth]{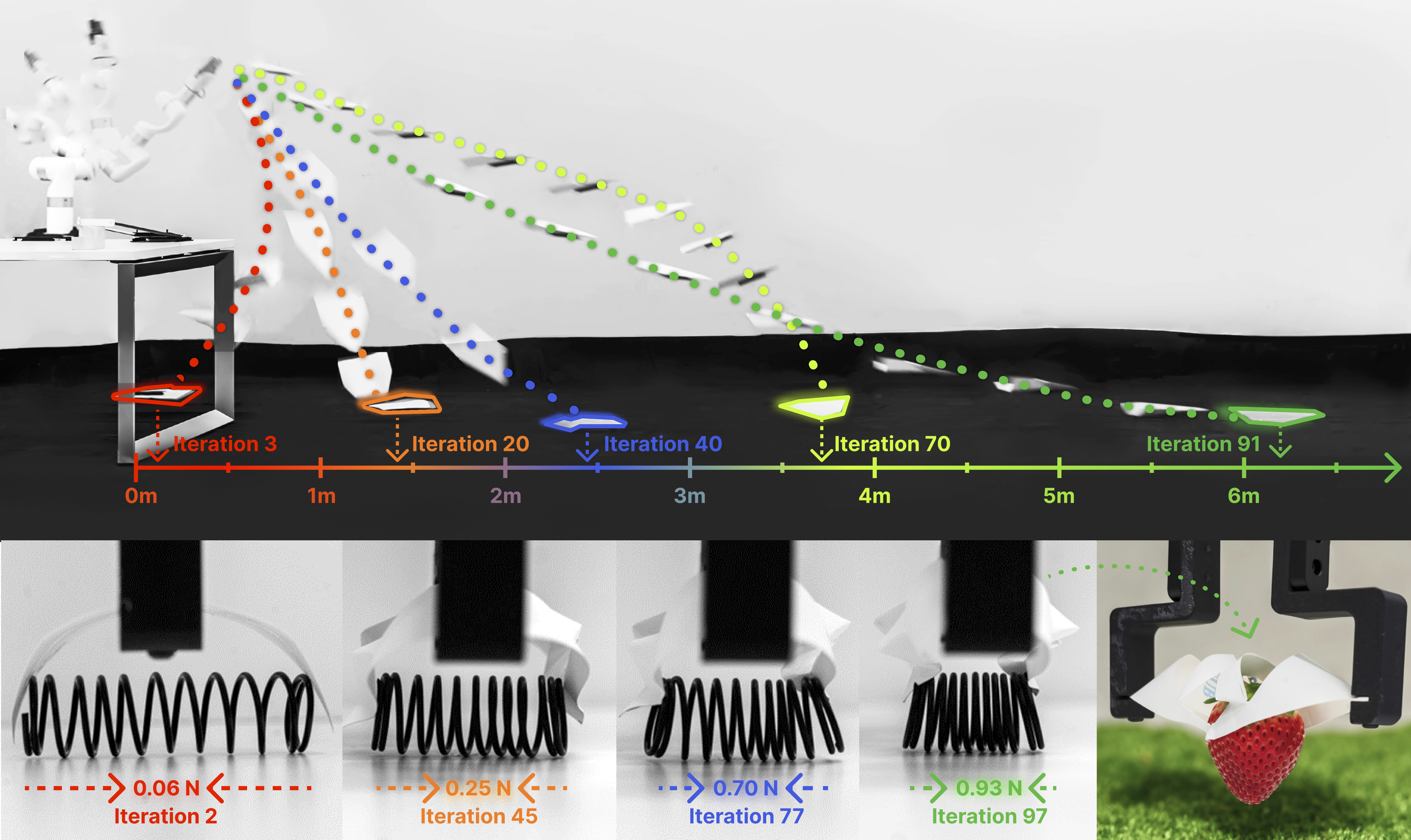}
                \captionof{figure}{
                    \textbf{
                        Self-supervised Design of Paper Tools.
                    }
                    Through trial-and-error, PaperBot autonomously learns how to design and use paper tools directly in the real world.
                    Given only 100 trials ($\approx$ 3 hours), our fully autonomous system discovers a paper airplane folding and throwing strategy that flew further than the best human design after the same number of trials (top), and learns how to cut and actuate a kirigami gripper that exerts 0.93N of force (bottom), equivalent to the weight of over four strawberries. For our system's final design, please see Fig.~\ref{fig:parameterization}.
                }
                \label{fig:teaser}
            \end{center}
        }]

\begin{abstract}
    Paper is a recyclable\footnote{ \vspace{-0.8cm} All the paper used during experiments was properly recycled.}, affordable, and widely accessible material, making it a popular medium for building  practical tools. 
    Traditional tool design either relies on simulation or physical analysis, which is often inaccurate and time-consuming.
    In this paper, we propose PaperBot, a framework that directly learns to design a tool out of paper then use it in the real world. We demonstrated the effectiveness and efficiency of PaperBot on two tool design tasks: a) learning to fold and throw paper airplanes for maximum travel distance, and b) learning to cut paper into grippers that exert maximum gripping force.
    We present a self-supervised learning framework that performs cutting, sequential-folding, and dynamic throwing and actuation actions to discover and optimize the design of paper tools.
    We deploy our system to a real-world bi-manual robotic system to solve challenging design tasks involving aerodynamics (paper airplane), friction and deformation (paper gripper) that are difficult to simulate accurately and efficiently with  traditional tool design approaches.
\end{abstract}

\IEEEpeerreviewmaketitle




\section{Introduction}

If you wanted to fold this paper (yes, the sheet you are reading) into an airplane that travels a maximum distance, how would you do it? By experimenting with paper as a child, you have learned multiple ways to cut and fold objects out of paper. In this work, we aim to automate this iterative design, deploy, and improve loop with machine learning methods in order to create real-world tools using paper (see Fig.~\ref{fig:teaser}). 
Paper is an attractive medium for design because it is an affordable, recyclable, and versatile commodity that can be repurposed and customized for different functions. Society has made many different tools out of paper -- ranging from filters and fans to packaging boxes and bowls, and it is beneficial to the environment to make more.  Robots could also create and use its own paper tools for many applications, for example in medical settings~\cite{cianchetti2018biomedical} where the cost of a paper end-effector allows it to be discarded after use or in logistical settings where paper vessels can be customized to each object.

Prior work in robotics has largely studied tool design in the context of simulation \cite{liu2023learning,he2023morph,taylor2019optimal,xu2021multi,kodnongbua2023computational}, where a reinforcement learning model learns the design of hardware jointly with a policy to use the tool.
Despite their impressive results, these prior works typically assume an abundance of data, which restricts their application to efficiently simulatable physical phenomena (i.e, rigid objects). 
In contrast, paper are deformable, thin-shell objects with relatively low elasticity and high plasticity.
Further, its applications typically involve interaction with fluids and air, which are expensive to simulate accurately.
Instead of scaling up high-quality synthetic data, how else could we approach automatic paper tool design?

In this paper, we introduce PaperBot, a robot system that autonomously performs experiments in the real world in order to learn the design of paper tools with different functions.
Given a reward function (such as travel distance or gripping force), the system learns a series of cuts or folds in order to create a paper tool that optimizes a reward. 
Our system finds these solutions through experimentation and learning:
\begin{itemize}
    \item \textbf{An automation pipeline that creates tools from sampled design parameters.}
          We build a system composed of bi-manual robot arms and several paper-related machines including a Cricut machine to perform a sequence of folding and cutting actions on paper. These actions are composed of a set of learnable parameters.

    \item \textbf{Automatic reward evaluation.}
          We use various sensors, including load cells and RGBD cameras to automatically evaluate the effectiveness of a tool created from the aforementioned automated pipeline and generate a reward.

    \item \textbf{An optimization framework that learns to optimize the design from trial-and-error.}
          We use a neural network as a surrogate model to learn the non-linear correlation between design parameters and reward. After training, optimal design parameters based on prior experimental data are searched in the surrogate model's parameter space. We adopt an epsilon-greedy strategy to search for the optimal parameters.
\end{itemize}

\begin{figure}[t]
    \centering
    \includegraphics[width=\linewidth]{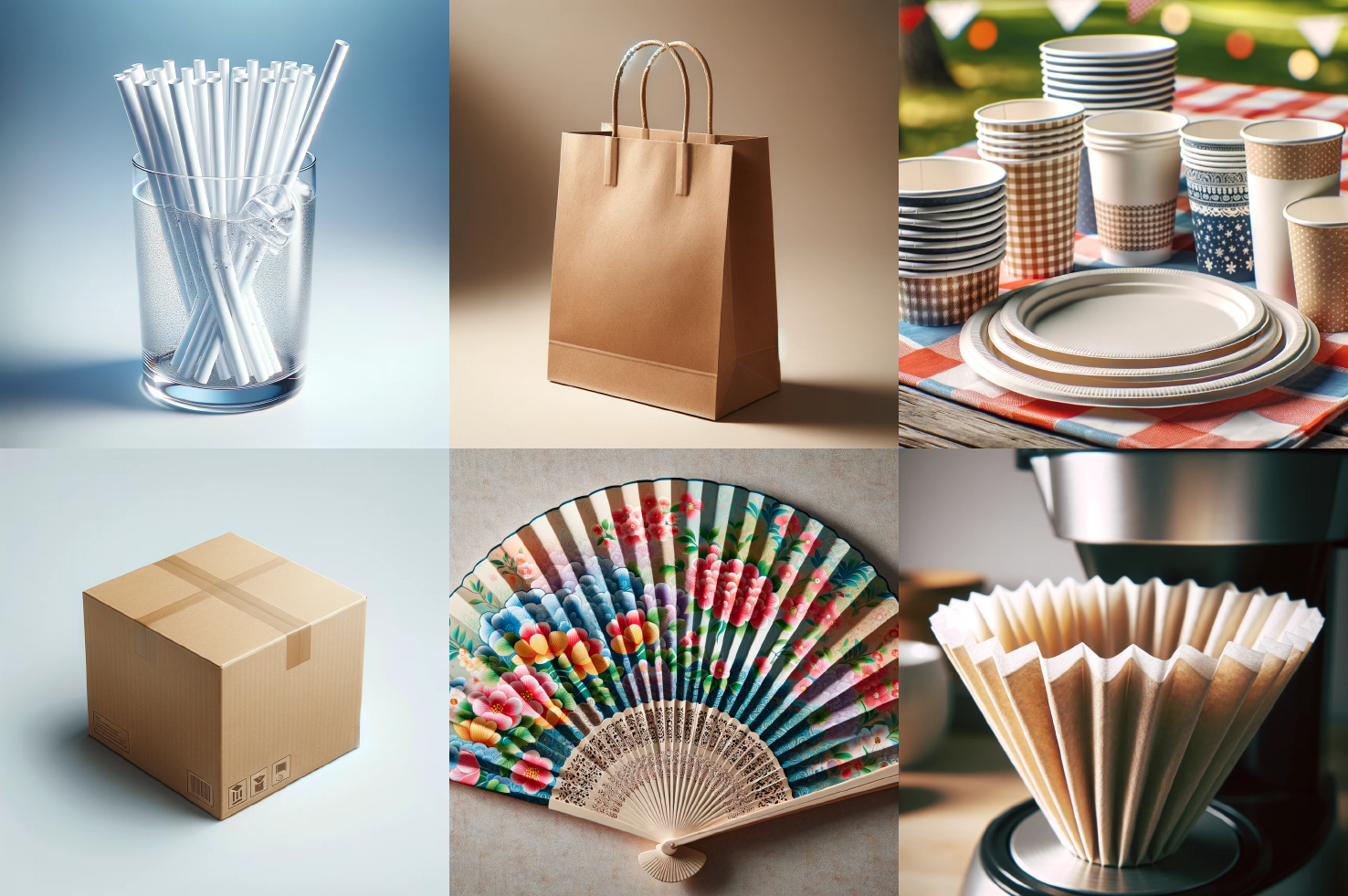}
    \caption{\textbf{Paper Tools.} Paper is an affordable and versatile medium for constructing a variety of different paper tools.}
    \label{fig:paper_tool}
\end{figure}

The optimization of the paper tool is driven by real-world feedback and operates entirely in the physical world without simulation and prior knowledge such as physical or material properties, making it generalizable to different tasks. Our approach learns strong designs with just 100 trials, requiring less than 3 hours for both tasks we evaluate. Experiments show that our learned design significantly outperforms baselines, including state-of-the-art methods based on evolutionary algorithms.

A key advantage of PaperBot is its ability to automatically adapt to new situations. In the real world, we often want customized tools for specific task definitions, which are expensive and time consuming to create if we need to hand-design every customization. PaperBot provides a viable way to automate this customization process. Our experiments show that the surrogate model trained on one task can quickly adapt to different reward definitions. 

The primary contribution of this paper is an approach for real-world tool design using paper as a material, and the rest of the paper will present and analyze this system in detail. Section~\ref{related} provides a brief overview of related work. Section~\ref{approach} introduces the approach to solve the design tasks of paper airplanes and kirigami grippers. Section~\ref{evaluation}  provides the experimental analysis of the system in terms of efficiency of the learning algorithm and the effectiveness of the designed tools. We believe the ability to perform real world tool design will significantly improve robot's ability to adapt and solve various real world tasks.

\begin{figure*}[t]
\includegraphics[width=\linewidth]{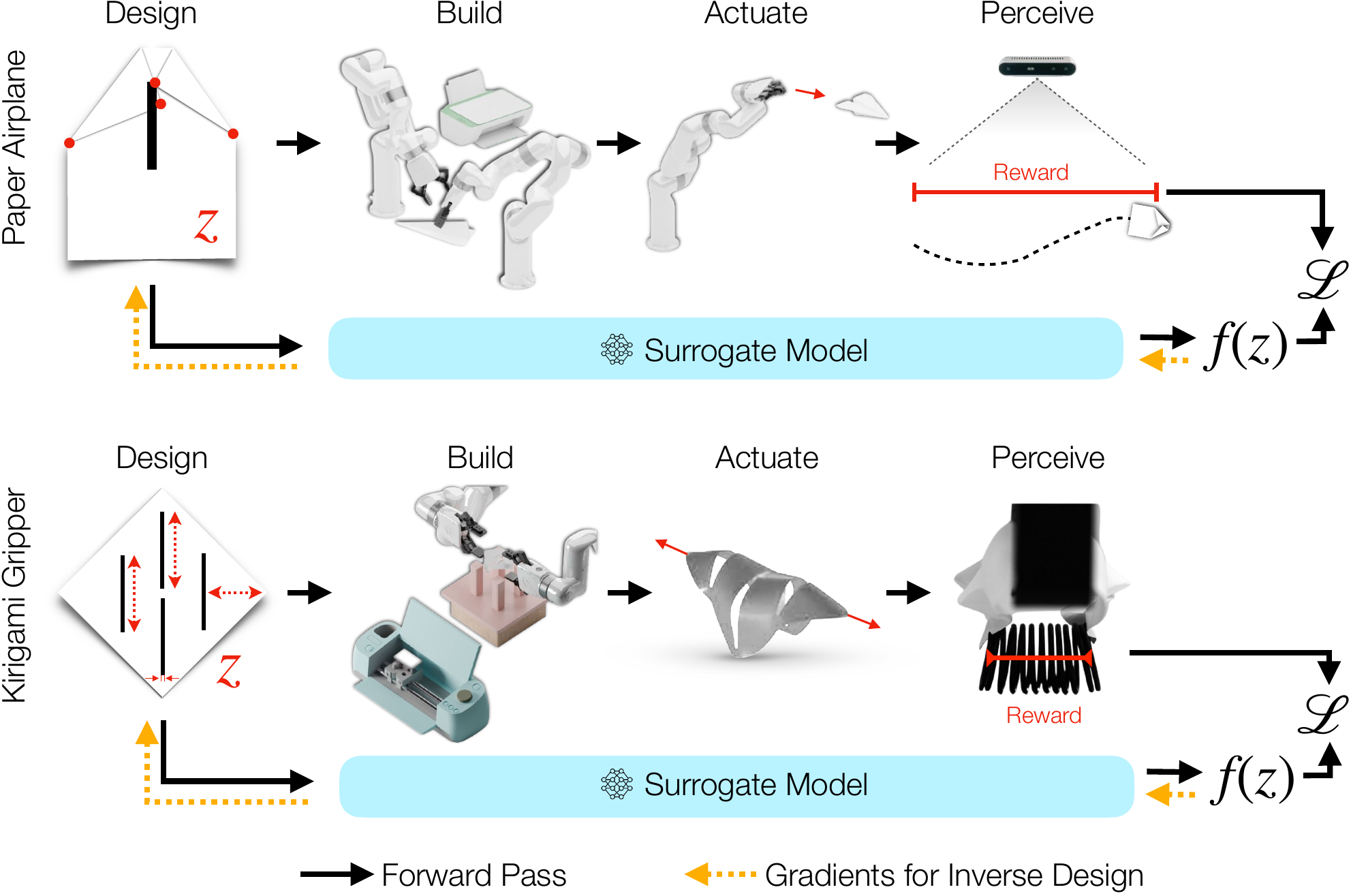}
    \caption{\textbf{Approach Overview.} Our framework samples paper designs for a tool, builds them, actuates a robot to perform on a task with the tool, and perceives its performance. By learning a surrogate model to predict the utility of a design, we obtain a differentiable model that allows us to solve inverse design tasks with gradient-based optimization. The above figure shows how our framework applies to two design tasks of paper airplanes (top) and kirigami grippers (bottom).}
    \label{fig:method}
\end{figure*}

\begin{figure*}[t]
    \centering
    \includegraphics[width=\textwidth]{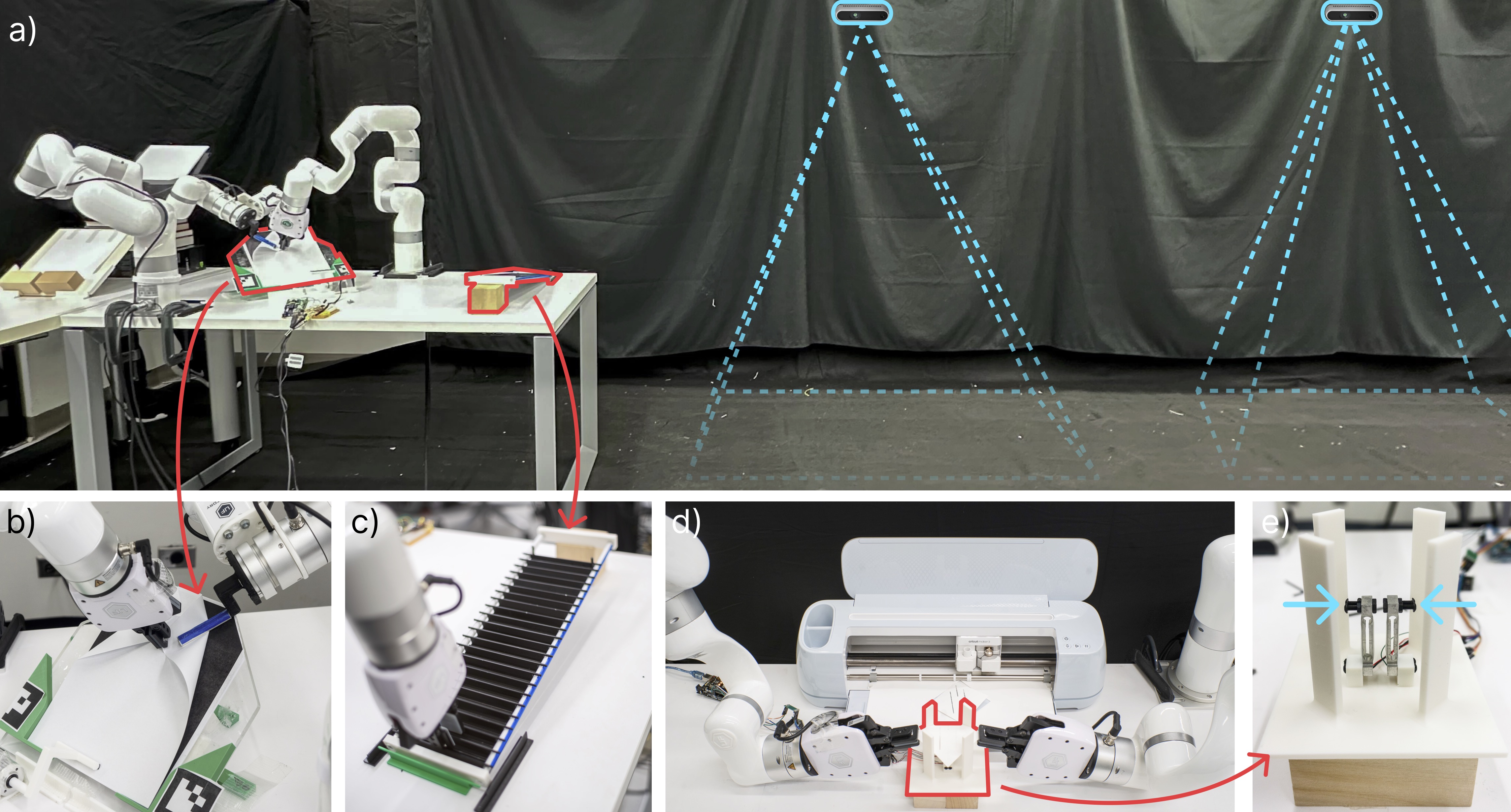}
    \caption{\textbf{System Setup.} We visualize the workspace of PaperBot while designing paper airplanes (\textbf{a}-\textbf{c}) and kirigami grippers (\textbf{d}-\textbf{e}). (\textbf{a}) shows the full workspace, including two RealSense D435i cameras to measure travel distance over the runway. (\textbf{b}) shows two xArm7s folding a paper airplane. (\textbf{c}) shows the sticky holders. (\textbf{d}) shows the xArm 7s retrieving a gripper cut by the Cricut Maker 3 and actuating it. (\textbf{e}) shows load cells that we use to measure the force of the candidate gripper design.}
    \label{fig:system}
\end{figure*}



\section{Related Work} \label{related}

\subsection{Origami-Inspired Robot and Tool Design}


The use of paper as a material in tool and robotic design has been explored in various fields. For instance, altering paper structures through cuts has been shown to modify mechanical properties such as stiffness~\cite{hwang2018tunable} and resonant frequency~\cite{yan2022cut}. Notably, these flat, 2D sheets can transform into complex 3D shapes, offering diverse design possibilities~\cite{isobe2016initial, choi2019programming, liu2019invariant}. For example, kirigami structures have been applied in optical tracking and enhancing solar panel power generation by controlling their deformation~\cite{lamoureux2015dynamic}. Similarly, origami principles have enabled the creation of transformable wheels capable of bearing significant loads~\cite{lee2021high}. Moreover, these deformable structures can imitate biological movements, such as crawling~\cite{rafsanjani2018kirigami} and aquatic swimming~\cite{ze2022spinning}. In more conventional applications, paper has been made into robotic manipulators~\cite{kim2018origami, jeong2018design, wu2021stretchable} and grippers~\cite{li2019vacuum, hong2022boundary, hong2023angle, buzzatto2023soft, kang2023grasping} for specialized object handling.

Previous studies have underscored the potential of paper-based designs, yet they typically require intricate human-driven analysis for optimization. This is attributed to the inherent challenge in simulating complex deformations and multiple buckling modes of thin-shell structures, making accurate predictions difficult~\cite{qiao2020elastic, yadav2021nondestructive, padhye2022isogeometric, narain2013folding}. The complexity of accurately predicting these behaviors necessitates advanced analysis and expertise.

\subsection{Hardware-Software Co-Design}


Real-world robots are often limited by their own morphologies as well as other physical constraints, posing many challenges when attempting to accomplish a task with policy learning alone. Therefore in recent years, hardware-software co-design has emerged as a field that studies how hardware design can be jointly learned with policy to create better tools for accomplishing certain tasks.

A line of work has explored solving manipulation tasks relying on simulation environments~\cite{xu2021multi,kodnongbua2023computational}, including~\cite{taylor2019optimal} that optimizes the shape of end-effector and motion jointly for dynamic planar manipulation. Other studies \cite{zhao2020robogrammar,xu2021multi, hu2022modular,whitman2020modular,hu2023glso,wang2023diffusebot} have investigated the hardware-software co-design problem using modular locomotion robots with data-driven approaches. With the rise of deep reinforcement learning and GPU-based large-scale simulation, a line of work~\cite{schaff2019jointly,chen2020hardware,liu2023learning} has focused on learning "hardware-as-policy" in simulated environments.
These works highlight the importance of integrating hardware and software in the design process. However, all prior works have approached the problem in a simulated environment, which is often hand-crafted for a specific application and difficult to tune accurately. In this work, we explore the possibility of learning hardware-software co-design directly in the real world by using paper as a material.

\subsection{Dynamic Manipulation of Deformable Objects}
The manipulation of deformable objects, such as paper, has been studied in the robotics and sciences community. The two main approaches to learning to manipulate an object are: model-based and model-free. In model-based works, a dynamics model of the real-world object enables the agent to simulate the outcomes of various forces acting upon the object and use this model to choose the next action. This dynamics model is either learned through particle-based dynamics network \cite{shi2022robocraft, shi2023robocook}, or via a differentiable simulator \cite{lin2022diffskill, huang2021plasticinelab}. Such approaches either require large amounts of training data due to the requirement of explicitly modeling dynamics, or assume good simulators. 

In model-free approaches, the relationship between input actions and output states is learned through interaction, with the dynamics of the interaction implicitly modeled. The concept of transporter networks, crucial in the manipulation of complex deformable objects, is investigated in \cite{seita2021learning, seita2023learning}. In dynamic manipulation \cite{ha2021flingbot,canberk2023cloth,chi2022irp,xu2022dextairity}, the action space is parameterized for efficiently learning a mapping to the goal state. PaperBot simultaneously learns to design \textit{and} dynamically manipulate a deformable object to achieve a goal state with a neural surrogate model to learn the non-linear correlation between design and manipulation parameters and reward through experimentation, then performing inverse design.

\section{Approach} \label{approach}

\begin{figure*}[t]
    \centering
    \includegraphics[width=\linewidth]{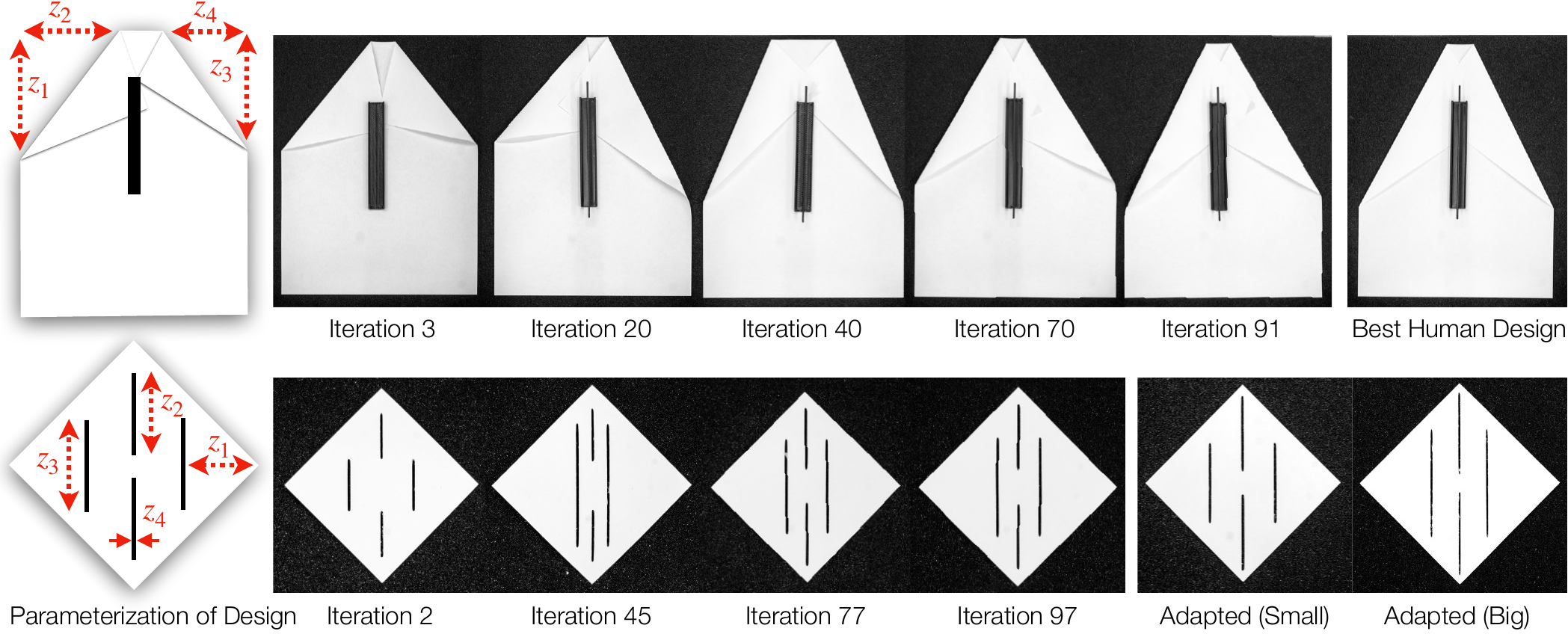}
    \caption{\textbf{Parameterization and Learning Process.} Left shows the parameterization of the tools. Right shows the intermediate designs during the learning process as well as the best human-fold plane and adapted gripper designs for smaller and larger objects. These iterations correspond to the teaser figure.  See Fig.~\ref{fig:teaser} for the actual performance of these designs.}
    \label{fig:parameterization}
\end{figure*}

\subsection{Inverse Design Framework}

PaperBot is a self-supervised framework for learning to design objects out of paper. Our approach performs real-world experiments with paper (cutting, folding) and automatically measures its performance on a task with a perception system. Given a parameterization of the design (such as where to cut or fold), we want to find a design $z \in \mathcal{Z} \subset \mathbb{R}^D$ that maximizes the reward function:
\begin{align}
    z^* = \argmax_z \; R(z)
    \label{eqn:optimal}
\end{align}
where the reward $R(z) \in \mathbb{R}$ is a perceivable quantity, such as travel distance or amount of force and $\mathcal{Z}$ is a bounded design parameter space. In real-world settings, calculating the reward $R$ requires building the paper tool, actuating a robot to use it, and measuring the reward with a perception system, i.e.\ $R(z) = \textrm{Perceive}(\textrm{Actuate}(\textrm{Build}(z), z))$. This process is neither convex nor differentiable, which makes optimizing the design challenging.

Our framework learns a differentiable surrogate model $f_\theta(z_t)$, represented by a neural network parameterized by $\theta$, that is trained to predict the reward of a candidate design $z_t$:
\begin{align}
    \min_\theta \; \mathbb{E}_{z_t} \left[
      \mathcal{L}\left(
        f_\theta(z_t), r_t
      \right)
    \right] \quad \textrm{for} \quad r_t = R(z_t)
    \label{eqn:learning}
\end{align}
where $z_t$ is the design configuration sampled at iteration $t$ and $\mathcal{L}$ is the loss function.
When sampling a new $z_{t+1}$, we use a $\epsilon$-greedy sampling strategy in order balance exploration and exploitation in the design space:
\begin{align}
    z_{t+1} = \begin{cases}
        \argmax_z \; f_\theta(z) & \textrm{with probability } 1 - \epsilon \\
        z \sim U & \textrm{with probability } \epsilon
    \end{cases}
\end{align}
where $U$ is a uniform distribution over $\mathcal{Z}$.

After a sufficient number of trials (we found 100 trials to be reasonable), the final design can be estimated with gradient descent by solving:
\begin{align}
    \hat{z} = \argmax_z \; f_\theta(z)
    \label{eqn:design-objective}
\end{align}
which approximates the optimal design $z^*$ in Equation \ref{eqn:optimal}.

In this work, we instantiate this framework on two paper design tasks: building paper airplanes that maximize their travel distance, and building kirigami grippers that exert a desired gripping force. In the remainder of this section, we describe how we apply this framework to these two tasks.




\subsection{Paper Airplane} \label{method:airplane}

We first consider the design of a paper airplane that travels a maximum distance. Prior works have explored similar tasks~\cite{obayashi2023automation,YouTube_2015} for different purposes, but here we use it as a motivating task for paper tool design as it involves a sequence of folding and throwing action as well as the interaction between deformable object and air, which is notably difficult to simulate.

\textbf{Task Configuration.}
The objective of the task is to fold and throw a paper airplane that maximizes the flying distance. The workspace is a 9m x 1.5m runway with curtains and wall on the side to prevent plane from deviating. A carpet was placed on the ground to prevent a paper airplane from sliding upon landing. We use the letter-size printer paper (216mm$\times$279mm) with 75$g/mm^2$ as material. The distance traveled by the plane is monitored by two RealSense D435i cameras mounted on the ceiling looking down with enough field of view to cover the entire runway.

\textbf{Design and Action Space.} The parameter space of the task is composed of 4 plane design parameters as described in Fig.~\ref{fig:parameterization}. We also learn 1 dynamic manipulation parameter of the throwing motion. Parameters 1 and 2 define the left fold position and orientation and parameters 3 and 4 define the right fold position and orientation. Parameters 5 is the throwing angle of the robot arm defined by when the gripper is released while the robot arm performs the throwing motion.

\begin{algorithm}[t]
\caption{$\epsilon$-Greedy Tool Design Optimization}
\label{algorithm}
\SetAlgoLined
\DontPrintSemicolon
\KwIn{Neural network $f$ with parameters $\theta$ and initialization $\sigma$, automation pipeline $R$, explore probability $\epsilon$, loss function $\mathcal{L}$, number of total trials $T$, optimization step sizes $\lambda_z, \lambda_\theta$, number of iterations $N$, $M$}
\KwOut{Estimated design parameters $\hat{z}$}
\For{$i = 1,\ldots,T$}{
    \eIf{\KwSty{rand()} $\leq$ $\epsilon$}{
        $z_0$ $\leftarrow$ randomly sampled design parameters\;
        $\{z\}_i = \{z\}_{i-1} \cup z_0$\;
        $\{r\}_i = \{r\}_{i-1} \cup R(z_0)$\;
    }{
        $\theta_0 \sim \mathcal{N}(0, \sigma)$ \;
        \tcp{Fitting surrogate model with data}
        \For{$j = 1,\ldots,N$} 
        {
            $B \leftarrow $  sample mini-batch from $\{z\}_i$, $\{r\}_i$\;
            $L \leftarrow \sum_{z,r \in B} \; \mathcal{L}\left(f_{\theta_{j-1}}(z), r\right)$ \;
            $\theta_j \leftarrow \theta_{j-1} - \lambda_\theta \frac{\partial}{\partial \theta_{j-1}} L$\;
        }
        \tcp{Inverse design}
        $z_0$ $\leftarrow$ randomly sampled design parameters\;
        \For{$j = 1, \ldots, M$}{
            $z_j \leftarrow z_{j-1} + \lambda_z \frac{\partial }{\partial z_{j-1}} f_{\theta_{N}}(z_{j-1})$\;
        }
        $\{z\}_i = \{z\}_{i-1} \cup z_{M}$\;
        $\{r\}_i = \{r\}_{i-1} \cup R(z_{M})$\;
    }
}
$t^* \leftarrow \argmax_{} {\{r\}}_T$\;
\Return{${z}_{t^*}$}\;
\end{algorithm}

\begin{figure}[t]
    \centering
    \subfloat[\centering Paper Airplane]{{\includegraphics[width=0.8\linewidth]{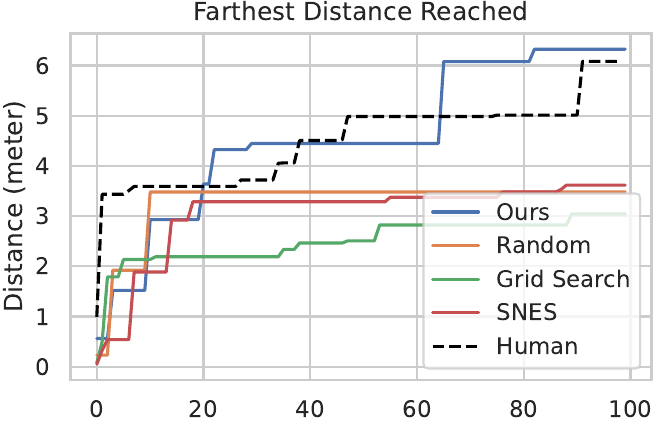}} } \label{fig:distance}
    \subfloat[\centering Kirigami Gripper]{{\includegraphics[width=0.85\linewidth]{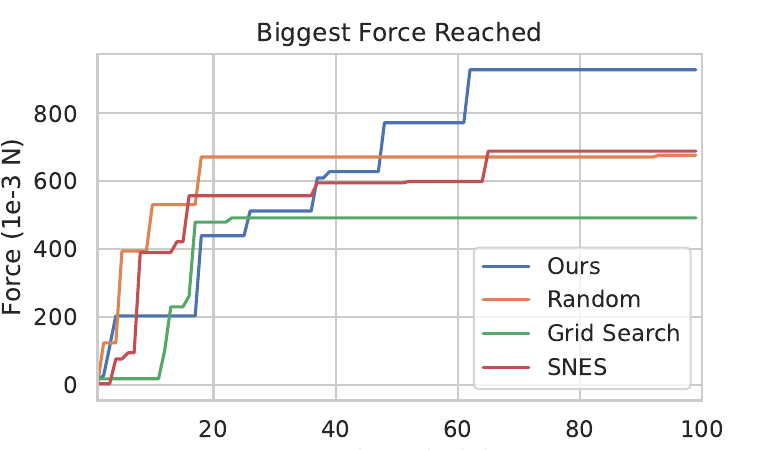}} }\label{fig:force}
    \caption{\textbf{Performance} of paper airplanes (top) and kirigami grippers (bottom) per optimization iteration,  designed by our method (blue) versus baselines.}
    \label{fig:quantitative}
    \vspace{0.5cm}
\end{figure}

\textbf{Automation.}
We automate paper airplane folding with two xArm 7 robot arms, one equipped with a parallel gripper and the other equipped with a 3D-printed soft presser tool and a suction cup. During folding, the corner of the paper is grasped by the parallel gripper and moved to the parameterized position, and the other arm will press down along the crease. The soft design of the presser tool exempts us from impedance control and did not fail once during more than a thousand trials we have run. To robustly load a single page of paper, we used a cheap inkjet printer that prints a blank page. To simplify the problem, after two folds, a 3D printed holder with pre-applied double-sided tape at the bottom is used to stick to the paper as a grasping point for throwing.

The pipeline for folding and throwing a paper airplane is fully automated without any human intervention. Each trial takes around 100 seconds, allowing 100 trials to be performed within 3 hours.

\subsection{Kirigami Gripper}  \label{method:gripper}

Kirigami grippers are inspired by the traditional art of paper cutting and folding, and they have recently been shown to be a novel approach for grasping in robotics~\cite{kirigami2021}. Constructed from a single flat sheet, these grippers are not only cost-effective and lightweight but also environmentally friendly, making them ideal for a variety of object grasping tasks in dynamic and unstructured environments~\cite{hong2022boundary, hong2023angle, buzzatto2023soft, kang2023grasping}. Nevertheless, the analysis and simulation of kirigami grippers pose significant challenges due to their complex behavior, including multiple buckling modes and reconfigurations upon contact with objects~\cite{chaudhary2023geometric, buzzatto2022soft}. This complexity requires sophisticated analysis and human expertise for kirigami gripper design in current methodologies. To resolve this, we propose to directly learn a surrogate model through experimentation in the real world and invert it for design.

\begin{figure*}[t!]
    \centering
    \includegraphics[width=\textwidth]{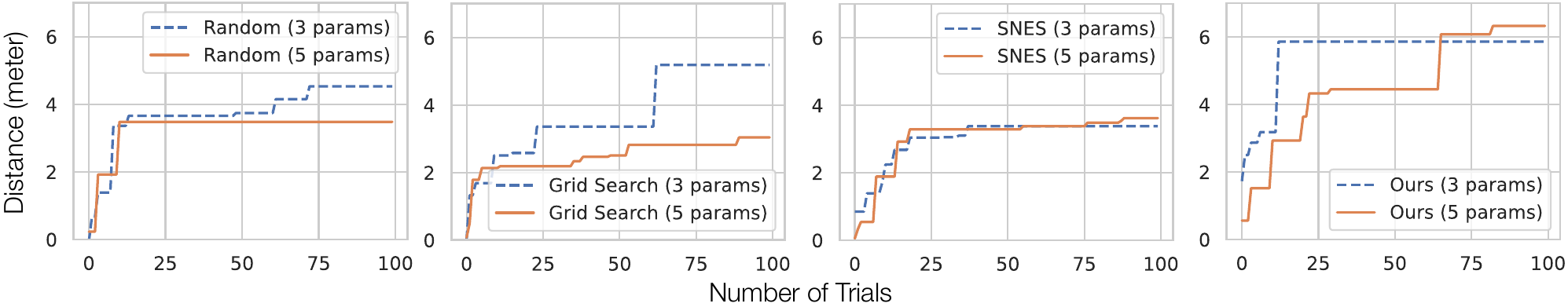}
    \caption{\textbf{Exploring Larger Design Spaces.} When the paper airplane design space increases from 3 to 5, random and grid search performs significantly worse.
    In contrast, our approach can efficiently navigate the larger design space to find better design/throwing parameters.
    Further, regardless of the design space, our approach significantly outperforms all baselines.}
    \label{fig:3v5}
\end{figure*}


\begin{figure}[t]
    \centering
    \includegraphics[width=\linewidth]{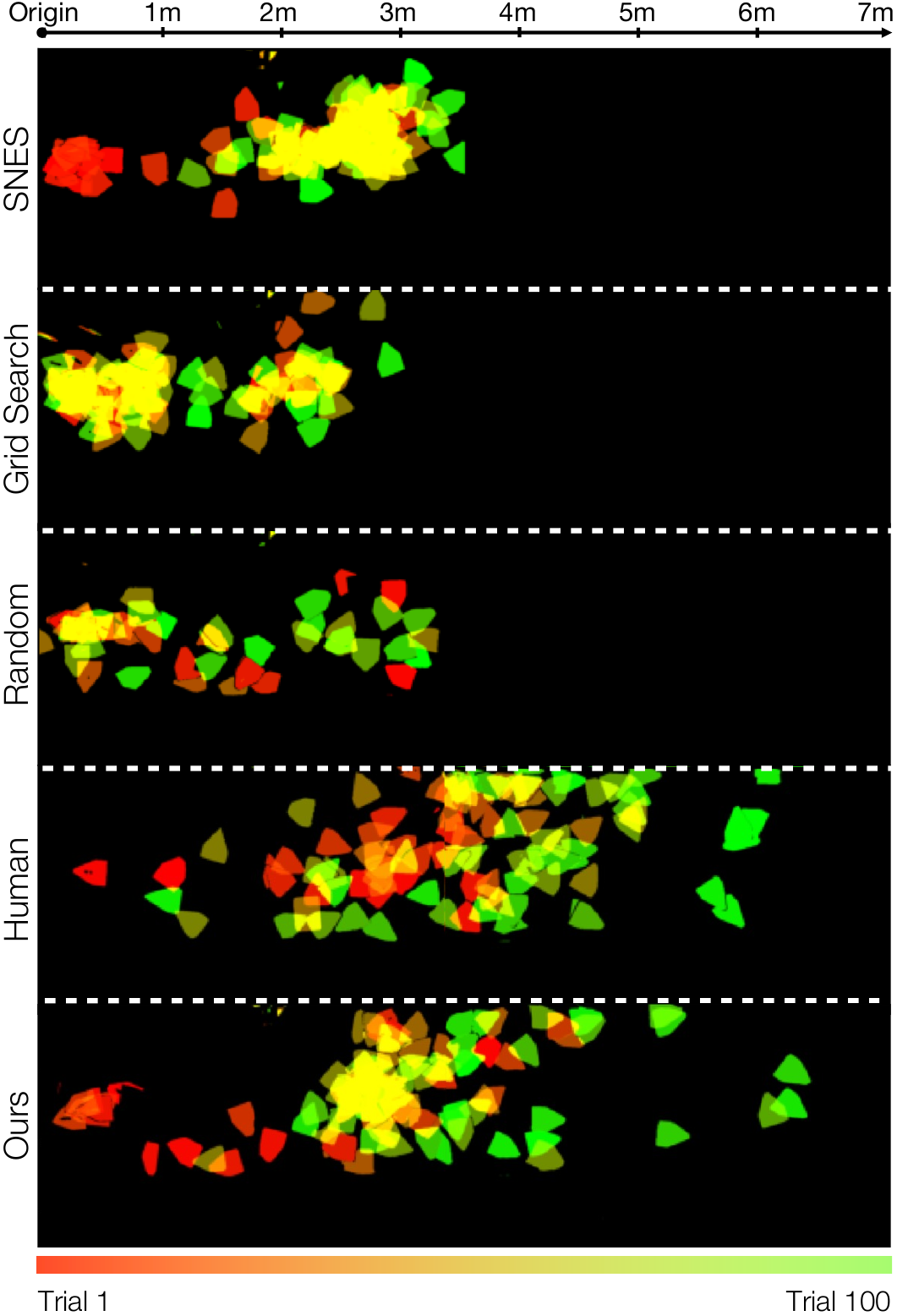}
    \caption{\textbf{Paper Airplanes Landing Positions.} 
    We visualize the top-down paper airplane mask over the course of training.
    Compared to SNES, PaperBot was able to better balance exploration and exploitation, escaping the local minima at around 2.5 meters, and eventually beating the human baseline.
    }
    \label{fig:distance_vis}
\end{figure}

\textbf{Task Configuration.}
The objective of this task is to design a gripper that exerts maximum gripping force for a given object geometry. The created gripper is placed on top of a measuring device with two parallel load cells measuring the horizontal gripping force exerted on each side of the gripper. A sequence of 240 measurements (80 Hz $\times$ 3s) is collected and applied with median filtering with a window size of 5. The peak value among a 10-step moving average is taken as the maximum force exerted by the Kirigami gripper. A sublimation paper (120 gsm) is used as the material\footnote{\href{https://a.co/d/2WK65p7}{https://a.co/d/2WK65p7}}.

\textbf{Design and Action Space.}
For simplicity and practicality, we keep the size of each gripper at 3x3 inches. We use 4 parameters to control the design of the gripper as described in Fig.~\ref{fig:parameterization}. This parameterization is functionally equivalent to the original design in~\cite{kirigami2021}, but more convenient for cutting. We limit the parameter space so that the cuts do not cross each other and do not get too close to the edge of the paper. Please see the detailed parameterization in Fig.~\ref{fig:parameterization}.

\textbf{Automation.}
We use a Cricut Maker 3 machine, which takes a PNG image of the designed gripper and cuts a sheet of paper into the exact shape. The cut gripper is placed to the force-measuring device shown in Fig.~\ref{fig:system}. Two xArm 7s with parallel grippers grasp the two opposite corners of the gripper and pull towards the opposite direction with 10mm/s speed for 30mm, during which both sides of the gripper press against the load cells. Different-sized 3D printed objects are mounted at the end of the load cells to mimic different-sized objects during adaptation experiments.

\subsection{Design Optimization}  \label{method:optim}

\begin{figure}[t]
    \centering
    \includegraphics[width=\linewidth]{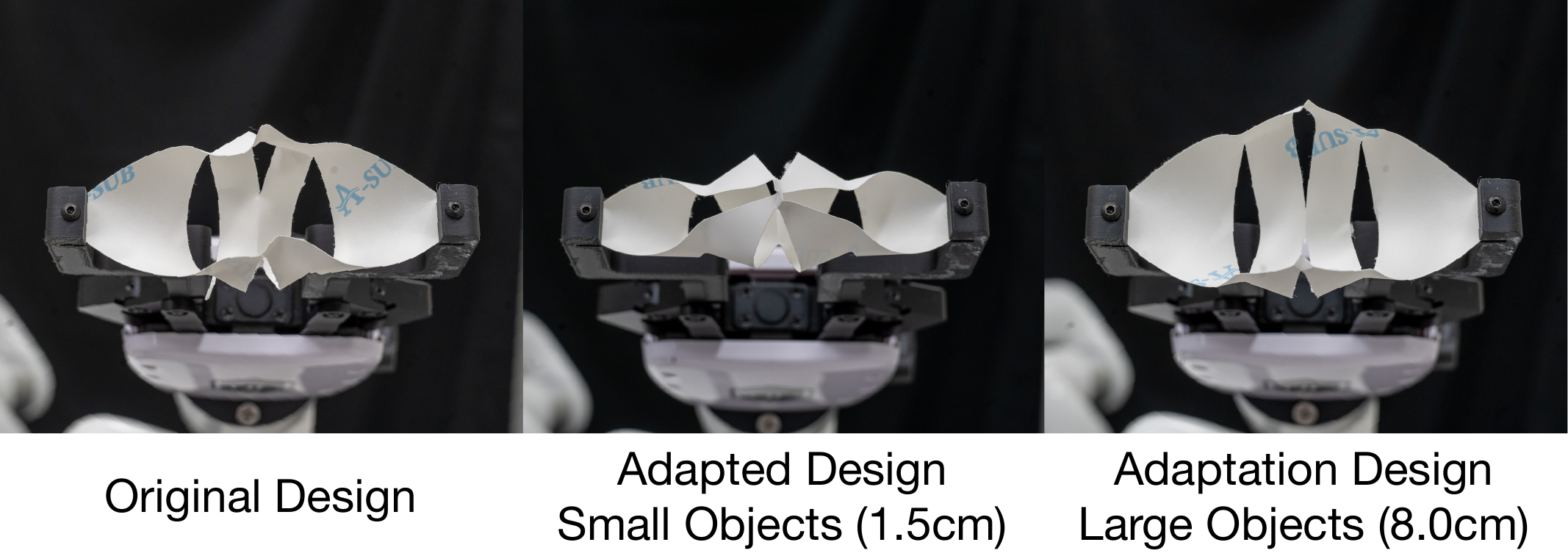}
    \caption{
    \textbf{Adapted Grippers Under Actuation.}
    Changing the distance between load cells changes where gripping force is measured, and thus, the optimal bending point for the gripper designs.
    Given only 50 adaptation trials, PaperBot discovered subtle design changes which significantly impacted the mechanical structure of the paper, leading to grippers tailored to different object sizes.
    }
    \label{fig:adaptation}
\end{figure}

After automating the construction and perception of the reward, we want to find a design $z$ that maximizes the reward function $R$ for a task. We simultaneously learn a surrogate model $f_\theta(z)$ and optimize designs against it.

\textbf{Learning the Surrogate Model.} We instantiate the surrogate model $f$ as a 4-layer MLP with an input dimension matching the corresponding parameter space, hidden dimensions of 512, and an scalar output for reward. Given a dataset of pairs of input design parameters and output reward, we train the network for $1000$ iterations with a batch size of 8. We use an AdamW optimizer with learning rate of 0.01 and weight decay of 0.1 to optimize for the Huber loss (smoothed L1 loss) between the predicted reward and ground truth.

Since we are operating in the real world, which is expensive in terms of time and resources, we cannot afford to massively randomly sample design parameters as training data for the surrogate model. For this, we adopt an $\epsilon$-greedy exploration strategy. At each iteration, there is an $\epsilon$ probability that we will perform random exploration, and otherwise, we perform a greedy search by finding the best design parameters defined by the neural surrogate model. Please see Algorithm~\ref{algorithm} for a detailed description.

\begin{figure*}[t]
    \centering
    \includegraphics[width=\textwidth]{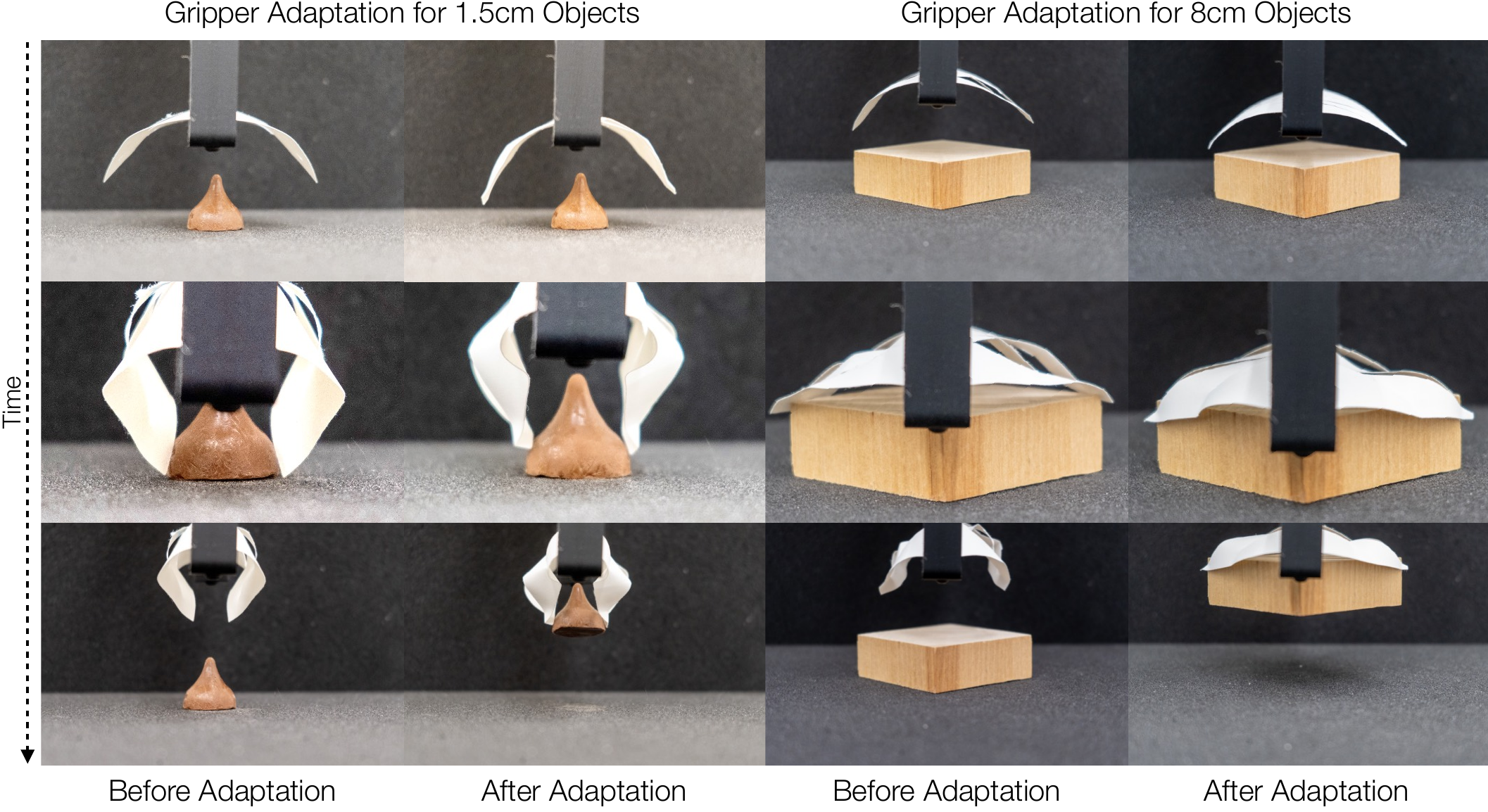}
    \caption{\textbf{Grasping with Adapted Gripper Design.} We compare the grasping process of our gripper design before adaptation and after adaptation for larger and smaller objects. Note that the ``before adaptation'' gripper is optimized for 5cm objects.}
    \label{fig:adaptation-qualitative}
\end{figure*}

\begin{figure*}[t]
    \centering
    \includegraphics[width=\textwidth]{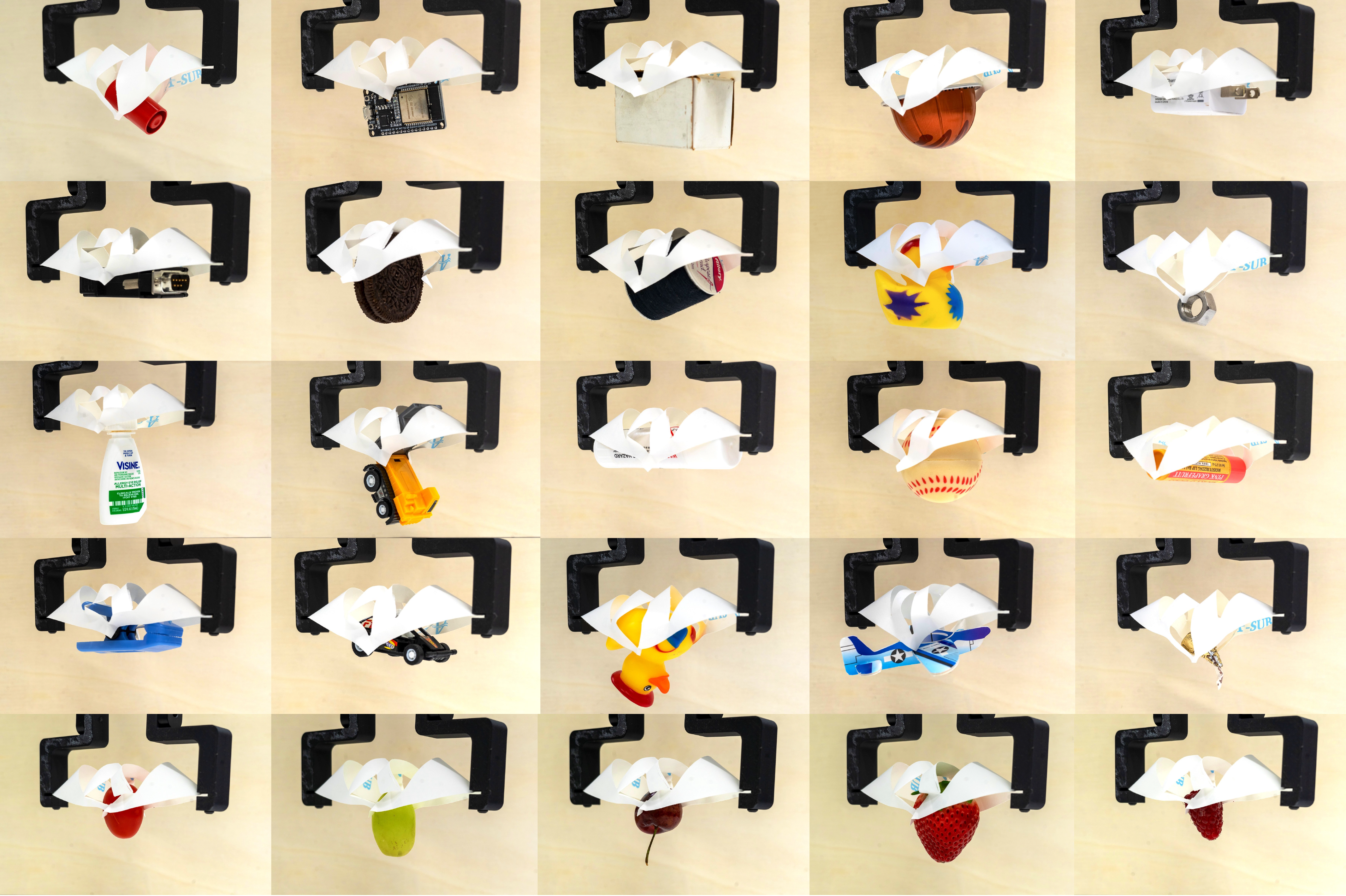}
    \caption{\textbf{Grasping of everyday objects using the optimized gripper design.} We used the grippers with a design optimized by our method to perform grasping of objects with various sizes, surface materials, weights, and rigidity, as well as fresh fruits as shown in the bottom row.}
    \label{fig:grasping}
\end{figure*}

\textbf{Inverse Design.}
Once the surrogate model is trained, we randomly sample a set of initial design parameters $z$ and perform gradient-descent on Equation \ref{eqn:design-objective}. During each iteration, we calculate $z \leftarrow z + \lambda \frac{\delta f_\theta(z)}{\delta z}$ for a scalar step size $\lambda \in \mathbb{R}$, which can be efficiently computed with back-propagation. The gradient contains information regarding which directions to change the design parameters to increase the reward. If the gradient update causes $z$ to deviate outside of physically feasible actions (such as cutting off the paper surface), we project $z$ back to the nearest feasible region.

\begin{table}[t]
\centering
\begin{tabular}{@{}lcc@{}}
\toprule
                        & \textbf{Small (1.5cm)} & \textbf{Large (8cm)} \\ 
\midrule
\textbf{Original} (optimized for 5cm)       & \(0.302 \pm 0.012\)    & \(0.055 \pm 0.027\)  \\
\textbf{Adapted Gripper} & \(0.442 \pm 0.080\)    & \(1.131 \pm 0.235\)   \\
\bottomrule
\end{tabular}
\caption{Force measurement of original gripper optimized for grasping 5cm-sized objects compared with adapted grippers optimized for different object sizes.}
\label{tab:adapt-quant}
\end{table}

Because the surrogate model training and inverse design combined take less than 2 seconds on an NVIDIA RTX 3090 GPU, far from being the bottleneck during experiments, we perform these two steps before each trial as detailed in Algorithm~\ref{algorithm}.

It is worth mentioning that evolutionary algorithms \cite{gomez2008accelerated,glasmachers2010exponential,schaul2011high,duan2016benchmarking,hansen2001completely} is another class of algorithms that also attempts to address the problem of gradient-free non-convex optimization. However, this class of methods is known to be sample-inefficient thus is not feasible in our real-world setting. We show the experimental comparison in section~\ref{exp:quantitative}.

\section{Evaluation} \label{evaluation}
The goal of this section is to systematically evaluate our system with both quantitative and qualitative analysis. In section~\ref{exp:quantitative}, we compare our method against several baselines to validate the effectiveness and efficiency of our optimization algorithm. In section~\ref{exp:size}, we show the effect of increasing the size of parameter space on ours and baseline methods. In section~\ref{exp:adapatation}, we show how our pre-trained model in one environment can adapt to a new environment with much less data. Finally, in section~\ref{exp:downstream}, we show applications of our optimized tool design.

\subsection{Quantitative Experiments} \label{exp:quantitative}
\textbf{Baselines.} For both tasks, we selected several baselines for comparison.
\begin{itemize}
    \item \textit{Grid Search:} A typical approach for parameter optimization is grid search. Given 100 trials as a budget, we used 5-5-4 for 3-parameter problems, 3-3-3-4 for 4-parameter problems, and 3-3-3-2-2 for 5-parameter problems.
    \item \textit{Random Sampling:} We perform random sampling of design parameters for each trial. Random and grid search provide a holistic view of the design problems' difficulty.
    \item \textit{SNES~\cite{snes}:} Evolutionary algorithms are common choices for solving gradient-free non-convex optimization problems. It has also been used in prior works for tool design based on simulation. We ran the algorithm with a population size of 5 for 20 iterations. We adopted the implementation provided by \href{https://evotorch.ai/}{EvoTorch}.
\end{itemize}

\textbf{Paper Airplane.}
In the paper airplane task, as shown in Fig.~\ref{fig:quantitative}, our approach significantly outperforms other baselines by a large margin, outperforming the second best baseline, SNES, by $87\%$.

To further understand the results, we visualize the landing positions of all folded paper airplanes during all experiment runs. This is collected by the ceiling-mounted RealSense cameras. The masks of planes of all trials are combined to visualize the distances traveled in Fig.~\ref{fig:distance_vis}. In both grid search and random exploration experiments, the majority of the folded paper airplanes landed near the origin, and overall show similar distance distribution. In SNES~\cite{snes}, most of the planes landed between 2.5-3.5 meters, while in comparison, our method quickly breaks out of that region and reaches further and further distances. During experiments, we observed that the evolutionary algorithm can quickly learn the throwing angle, which has the highest correlation with the travel distance and is relatively orthogonal to other design parameters. However, during the 100 trials, SNES struggled to find good folding parameters, which are necessary for folding a paper airplane that flies further than 3.5 meters. This explains what we see in the visualization and shows that our method is significantly more sample-efficient than evolutionary algorithm for this problem in a low-data setting.


\vspace{0.5cm}
\textbf{Kirigami Gripper.}
For the kirigami gripper design task, we adopted the same set of baselines. We optimize the 4 design parameters as shown in Fig.~\ref{fig:parameterization}. Similar to the paper airplane task, our method outperforms the second-best baseline -- SNES by a large margin of 36.4$\%$. Our best-designed gripper can exert a gripping force equivalent to the gravity of 92.8g, which is more than the weight of four strawberries.

\subsection{Size of Design Space} \label{exp:size}

Since real-world tools are complicated and contain many design parameters, we want to understand the scaling performance of different methods against the number of parameters. To investigate this, we perform an ablation study on the paper airplane task. We run our experiments to learn 3 and 5 parameters independently. For the 3 parameter experiments, we cut the two parameters governing the right fold, and replaced them with the parameters of the left fold, enforcing a symmetry of wing angles to every plane -- reducing the complexity of the design task.

As shown in figure~\ref{fig:3v5}, the performance of baseline methods drops significantly as the number of learnable parameters increases. All of these methods struggle to learn to fold symmetrically, which causes most of the folded planes to crash on the side. However, for our method, even though the rate of learning in the 5-parameter experiment is much slower than the 3-parameter one, it eventually found a better design. We observed that this is because our enforced symmetry does not guarantee the plane travel in a straight line due to both calibration precision and environmental influence (such as small airflow on the paper runway). When given 5 learnable parameters, our method can discover a slightly asymmetric design that counteracts the systematic asymmetry, allowing the plane to travel in a straight line to the furthest distance.

\subsection{Adaptation of Kirigami Gripper to Different Objects} \label{exp:adapatation}

Since the entire learning process happens in the real world and we do not assume prior knowledge about the physical world, our system is highly adaptive. In the kirigami gripper design task, we conduct experiments where we vary the distance between the load cells to mimic objects of different sizes. In these experiments, we initialized the parameters of the neural surrogate model with pretrained weights from previous experiments conducted on a different object size. With fewer experiments (50 trials), our method can quickly adapt its design for different object sizes.

Table~\ref{tab:adapt-quant} shows the force measurement of the original gripper vs. adapted gripper exerted on new objects. For the small object experiment, the adapted gripper exerts 46$\%$ more gripping force than before. For larger objects, the adapted gripper obtains a more than 20-fold increase in gripping force.

The effect of adaptation of gripper design can be qualitatively seen in Fig.~\ref{fig:adaptation}, where we mount different grippers on a parallel robot gripper and actuate with the same stretching distances. The difference in design changes the mechanical structure of the paper during its deformation, causing it to enclose objects of different sizes much better. This effect directly causes the success rate of grasping objects of different sizes as shown in Fig.~\ref{fig:adaptation-qualitative}, illustrating the importance of customization. These experiments showcase the advantage of our method to quickly adapt to different reward definitions and potentially different environmental conditions. It has also shown that prior knowledge obtained from a related task with a shared structure can be effectively transferred to improve sample efficiency.

\subsection{Grasping with Designed Gripper} \label{exp:downstream}

In Fig.~\ref{fig:grasping}, the robot performs grasping on a variety of objects with different sizes and geometry to showcase the effectiveness of the emergent gripper design. In each example, we mount the kirigami gripper on a normal robot parallel gripper using a 3D printer tool and perform a predefined primitive grasping action. These objects include small, high-density objects such as the metal nut, as well as large objects with a smooth slippery surface such as the box, highlighting that our optimized design can exert high gripping force for grasping. The bottom row shows our designed gripper grasping various fruits, which motivates paper as a material for making cheap, clean, recyclable, and customizable grippers for handling food.

\section{Conclusion}
\label{sec:conclusion}

Paper is an affordable and versatile medium that can be used to construct many different types of tools. We demonstrated PaperBot, a system that autonomously performs experiments to simultaneously learn the a) design and b) use of paper tools for different tasks. Without requiring a simulator, experiments show that our approach is able to learn to design, build, and use real-world paper tools in just a few hours. Our system is reproducible in any robotics research lab and we will open-source all software, hardware, models, and data.

\section*{Acknowledgments}

This research is based on work partially supported by NSF NRI Awards \#1925157 and \#2132519, and the Toyota Research Institute. S.S.\ is supported by the NSF Graduate Research Fellowship.


\bibliographystyle{plainnat}
\bibliography{references}

\begin{thebibliography}{55}
\providecommand{\natexlab}[1]{#1}
\providecommand{\url}[1]{\texttt{#1}}
\expandafter\ifx\csname urlstyle\endcsname\relax
  \providecommand{\doi}[1]{doi: #1}\else
  \providecommand{\doi}{doi: \begingroup \urlstyle{rm}\Url}\fi

\bibitem[Buzzatto et~al.(2022)Buzzatto, Shahmohammadi, Liang, Sanches, Matsunaga, Haraguchi, Mariyama, MacDonald, and Liarokapis]{buzzatto2022soft}
Joao Buzzatto, Mojtaba Shahmohammadi, Junbang Liang, Felipe Sanches, Saori Matsunaga, Rintaro Haraguchi, Toshisada Mariyama, Bruce MacDonald, and Minas Liarokapis.
\newblock Soft, multi-layer, disposable, kirigami based robotic grippers: On handling of delicate, contaminated, and everyday objects.
\newblock In \emph{2022 IEEE/RSJ International Conference on Intelligent Robots and Systems (IROS)}, pages 5440--5447. IEEE, 2022.

\bibitem[Buzzatto et~al.(2023)Buzzatto, Liang, Shahmohammadi, Matsunaga, Haraguchi, Mariyama, MacDonald, and Liarokapis]{buzzatto2023soft}
Joao Buzzatto, Junbang Liang, Mojtaba Shahmohammadi, Saori Matsunaga, Rintaro Haraguchi, Toshisada Mariyama, Bruce~A MacDonald, and Minas Liarokapis.
\newblock A soft, multi-layer, kirigami inspired robotic gripper with a compact, compression-based actuation system.
\newblock In \emph{2023 IEEE/RSJ International Conference on Intelligent Robots and Systems (IROS)}, pages 4488--4495. IEEE, 2023.

\bibitem[Canberk et~al.(2023)Canberk, Chi, Ha, Burchfiel, Cousineau, Feng, and Song]{canberk2023cloth}
Alper Canberk, Cheng Chi, Huy Ha, Benjamin Burchfiel, Eric Cousineau, Siyuan Feng, and Shuran Song.
\newblock Cloth funnels: Canonicalized-alignment for multi-purpose garment manipulation.
\newblock In \emph{2023 IEEE International Conference on Robotics and Automation (ICRA)}, pages 5872--5879. IEEE, 2023.

\bibitem[Chaudhary et~al.(2023)Chaudhary, Niu, Han, Lewicka, and Mahadevan]{chaudhary2023geometric}
G~Chaudhary, L~Niu, Q~Han, M~Lewicka, and L~Mahadevan.
\newblock Geometric mechanics of ordered and disordered kirigami.
\newblock \emph{Proceedings of the Royal Society A}, 479\penalty0 (2274):\penalty0 20220822, 2023.

\bibitem[Chen et~al.(2020)Chen, He, and Ciocarlie]{chen2020hardware}
Tianjian Chen, Zhanpeng He, and Matei Ciocarlie.
\newblock Hardware as policy: Mechanical and computational co-optimization using deep reinforcement learning.
\newblock \emph{arXiv preprint arXiv:2008.04460}, 2020.

\bibitem[Chi et~al.(2022)Chi, Burchfiel, Cousineau, Feng, and Song]{chi2022irp}
Cheng Chi, Benjamin Burchfiel, Eric Cousineau, Siyuan Feng, and Shuran Song.
\newblock Iterative residual policy for goal-conditioned dynamic manipulation of deformable objects.
\newblock In \emph{Proceedings of Robotics: Science and Systems (RSS)}, 2022.

\bibitem[Choi et~al.(2019)Choi, Dudte, and Mahadevan]{choi2019programming}
Gary~PT Choi, Levi~H Dudte, and Lakshminarayanan Mahadevan.
\newblock Programming shape using kirigami tessellations.
\newblock \emph{Nature materials}, 18\penalty0 (9):\penalty0 999--1004, 2019.

\bibitem[Cianchetti et~al.(2018)Cianchetti, Laschi, Menciassi, and Dario]{cianchetti2018biomedical}
Matteo Cianchetti, Cecilia Laschi, Arianna Menciassi, and Paolo Dario.
\newblock Biomedical applications of soft robotics.
\newblock \emph{Nature Reviews Materials}, 3\penalty0 (6):\penalty0 143--153, 2018.

\bibitem[Duan et~al.(2016)Duan, Chen, Houthooft, Schulman, and Abbeel]{duan2016benchmarking}
Yan Duan, Xi~Chen, Rein Houthooft, John Schulman, and Pieter Abbeel.
\newblock Benchmarking deep reinforcement learning for continuous control.
\newblock In \emph{International conference on machine learning}, pages 1329--1338. PMLR, 2016.

\bibitem[Glasmachers et~al.(2010)Glasmachers, Schaul, Yi, Wierstra, and Schmidhuber]{glasmachers2010exponential}
Tobias Glasmachers, Tom Schaul, Sun Yi, Daan Wierstra, and J{\"u}rgen Schmidhuber.
\newblock Exponential natural evolution strategies.
\newblock In \emph{Proceedings of the 12th annual conference on Genetic and evolutionary computation}, pages 393--400, 2010.

\bibitem[Gomez et~al.(2008)Gomez, Schmidhuber, Miikkulainen, and Mitchell]{gomez2008accelerated}
Faustino Gomez, J{\"u}rgen Schmidhuber, Risto Miikkulainen, and Melanie Mitchell.
\newblock Accelerated neural evolution through cooperatively coevolved synapses.
\newblock \emph{Journal of Machine Learning Research}, 9\penalty0 (5), 2008.

\bibitem[Ha and Song(2021)]{ha2021flingbot}
Huy Ha and Shuran Song.
\newblock Flingbot: The unreasonable effectiveness of dynamic manipulation for cloth unfolding.
\newblock In \emph{Conference on Robotic Learning (CoRL)}, 2021.

\bibitem[Hansen and Ostermeier(2001)]{hansen2001completely}
Nikolaus Hansen and Andreas Ostermeier.
\newblock Completely derandomized self-adaptation in evolution strategies.
\newblock \emph{Evolutionary computation}, 9\penalty0 (2):\penalty0 159--195, 2001.

\bibitem[He and Ciocarlie(2023)]{he2023morph}
Zhanpeng He and Matei Ciocarlie.
\newblock Morph: Design co-optimization with reinforcement learning via a differentiable hardware model proxy.
\newblock \emph{arXiv preprint arXiv:2309.17227}, 2023.

\bibitem[Hong et~al.(2022)Hong, Chi, Wu, Li, Zhu, and Yin]{hong2022boundary}
Yaoye Hong, Yinding Chi, Shuang Wu, Yanbin Li, Yong Zhu, and Jie Yin.
\newblock Boundary curvature guided programmable shape-morphing kirigami sheets.
\newblock \emph{Nature Communications}, 13\penalty0 (1):\penalty0 530, 2022.

\bibitem[Hong et~al.(2023)Hong, Zhao, Berman, Chi, Li, Huang, and Yin]{hong2023angle}
Yaoye Hong, Yao Zhao, Joseph Berman, Yinding Chi, Yanbin Li, He~Huang, and Jie Yin.
\newblock Angle-programmed tendril-like trajectories enable a multifunctional gripper with ultradelicacy, ultrastrength, and ultraprecision.
\newblock \emph{Nature Communications}, 14\penalty0 (1):\penalty0 4625, 2023.

\bibitem[Hu et~al.(2022)Hu, Whitman, Travers, and Choset]{hu2022modular}
Jiaheng Hu, Julian Whitman, Matthew Travers, and Howie Choset.
\newblock Modular robot design optimization with generative adversarial networks.
\newblock In \emph{2022 International Conference on Robotics and Automation (ICRA)}, pages 4282--4288. IEEE, 2022.

\bibitem[Hu et~al.(2023)Hu, Whitman, and Choset]{hu2023glso}
Jiaheng Hu, Julian Whitman, and Howie Choset.
\newblock Glso: Grammar-guided latent space optimization for sample-efficient robot design automation.
\newblock In \emph{Conference on Robot Learning}, pages 1321--1331. PMLR, 2023.

\bibitem[Huang et~al.(2021)Huang, Hu, Du, Zhou, Su, Tenenbaum, and Gan]{huang2021plasticinelab}
Zhiao Huang, Yuanming Hu, Tao Du, Siyuan Zhou, Hao Su, Joshua~B. Tenenbaum, and Chuang Gan.
\newblock Plasticinelab: A soft-body manipulation benchmark with differentiable physics, 2021.

\bibitem[Hwang and Bartlett(2018)]{hwang2018tunable}
Doh-Gyu Hwang and Michael~D Bartlett.
\newblock Tunable mechanical metamaterials through hybrid kirigami structures.
\newblock \emph{Scientific reports}, 8\penalty0 (1):\penalty0 3378, 2018.

\bibitem[Isobe and Okumura(2016)]{isobe2016initial}
Midori Isobe and Ko~Okumura.
\newblock Initial rigid response and softening transition of highly stretchable kirigami sheet materials.
\newblock \emph{Scientific reports}, 6\penalty0 (1):\penalty0 24758, 2016.

\bibitem[Jeong and Lee(2018)]{jeong2018design}
Donghwa Jeong and Kiju Lee.
\newblock Design and analysis of an origami-based three-finger manipulator.
\newblock \emph{Robotica}, 36\penalty0 (2):\penalty0 261--274, 2018.

\bibitem[Kang et~al.(2023)Kang, Kim, Lee, Kim, Lee, and Song]{kang2023grasping}
Gyeongji Kang, Young-Joo Kim, Sung-Jin Lee, Se~Kwon Kim, Dae-Young Lee, and Kahye Song.
\newblock Grasping through dynamic weaving with entangled closed loops.
\newblock \emph{Nature Communications}, 14\penalty0 (1):\penalty0 4633, 2023.

\bibitem[Kim et~al.(2018)Kim, Lee, Jung, and Cho]{kim2018origami}
Suk-Jun Kim, Dae-Young Lee, Gwang-Pil Jung, and Kyu-Jin Cho.
\newblock An origami-inspired, self-locking robotic arm that can be folded flat.
\newblock \emph{Science Robotics}, 3\penalty0 (16):\penalty0 eaar2915, 2018.

\bibitem[Kodnongbua et~al.(2023)Kodnongbua, Lou, Lipton, and Schulz]{kodnongbua2023computational}
Milin Kodnongbua, Ian Good~Yu Lou, Jeffrey Lipton, and Adriana Schulz.
\newblock Computational design of passive grippers.
\newblock \emph{arXiv preprint arXiv:2306.03174}, 2023.

\bibitem[Lamoureux et~al.(2015)Lamoureux, Lee, Shlian, Forrest, and Shtein]{lamoureux2015dynamic}
Aaron Lamoureux, Kyusang Lee, Matthew Shlian, Stephen~R Forrest, and Max Shtein.
\newblock Dynamic kirigami structures for integrated solar tracking.
\newblock \emph{Nature communications}, 6\penalty0 (1):\penalty0 8092, 2015.

\bibitem[Lee et~al.(2021)Lee, Kim, Sohn, Heo, and Cho]{lee2021high}
Dae-Young Lee, Jae-Kyeong Kim, Chang-Young Sohn, Jeong-Mu Heo, and Kyu-Jin Cho.
\newblock High--load capacity origami transformable wheel.
\newblock \emph{Science Robotics}, 6\penalty0 (53):\penalty0 eabe0201, 2021.

\bibitem[Li et~al.(2019)Li, Stampfli, and Xu]{li2019vacuum}
S~Li, JJ~Stampfli, and HJ~Xu.
\newblock A vacuum-driven origami “magic-ball” soft gripper” in 2019 international conference on robotics and automation (icra).
\newblock \emph{IEEE, New York}, pages 7401--7408, 2019.

\bibitem[Lin et~al.(2022)Lin, Huang, Li, Tenenbaum, Held, and Gan]{lin2022diffskill}
Xingyu Lin, Zhiao Huang, Yunzhu Li, Joshua~B. Tenenbaum, David Held, and Chuang Gan.
\newblock Diffskill: Skill abstraction from differentiable physics for deformable object manipulations with tools.
\newblock 2022.

\bibitem[Liu et~al.(2019)Liu, Tachi, and Paulino]{liu2019invariant}
Ke~Liu, Tomohiro Tachi, and Glaucio~H Paulino.
\newblock Invariant and smooth limit of discrete geometry folded from bistable origami leading to multistable metasurfaces.
\newblock \emph{Nature communications}, 10\penalty0 (1):\penalty0 4238, 2019.

\bibitem[Liu et~al.(2023)Liu, Tian, Guo, Liu, and Wu]{liu2023learning}
Ziang Liu, Stephen Tian, Michelle Guo, Karen Liu, and Jiajun Wu.
\newblock Learning to design and use tools for robotic manipulation.
\newblock In \emph{Conference on Robot Learning}, pages 887--905. PMLR, 2023.

\bibitem[Narain et~al.(2013)Narain, Pfaff, and O'Brien]{narain2013folding}
Rahul Narain, Tobias Pfaff, and James~F O'Brien.
\newblock Folding and crumpling adaptive sheets.
\newblock \emph{ACM Transactions on Graphics (TOG)}, 32\penalty0 (4):\penalty0 1--8, 2013.

\bibitem[Obayashi et~al.(2023)Obayashi, Junge, Ilić, and Hughes]{obayashi2023automation}
Nana Obayashi, Kai Junge, Stefan Ilić, and Josie Hughes.
\newblock Robotic automation and unsupervised cluster assisted modeling for solving the forward and reverse design problem of paper airplanes.
\newblock \emph{Scientific Reports}, 13\penalty0 (1), Mar 2023.
\newblock \doi{10.1038/s41598-023-31395-0}.

\bibitem[Padhye and Kalia(2022)]{padhye2022isogeometric}
Nikhil Padhye and Subodh Kalia.
\newblock Isogeometric analysis of elastic sheets exhibiting combined bending and stretching using dynamic relaxation.
\newblock \emph{arXiv preprint arXiv:2206.10406}, 2022.

\bibitem[Qiao et~al.(2020)Qiao, Liu, and Pasini]{qiao2020elastic}
Chuan Qiao, Lu~Liu, and Damiano Pasini.
\newblock Elastic thin shells with large axisymmetric imperfection: From bifurcation to snap-through buckling.
\newblock \emph{Journal of the Mechanics and Physics of Solids}, 141:\penalty0 103959, 2020.

\bibitem[Rafsanjani et~al.(2018)Rafsanjani, Zhang, Liu, Rubinstein, and Bertoldi]{rafsanjani2018kirigami}
Ahmad Rafsanjani, Yuerou Zhang, Bangyuan Liu, Shmuel~M Rubinstein, and Katia Bertoldi.
\newblock Kirigami skins make a simple soft actuator crawl.
\newblock \emph{Science Robotics}, 3\penalty0 (15):\penalty0 eaar7555, 2018.

\bibitem[Robotics(2015)]{YouTube_2015}
ABB Robotics, Sep 2015.
\newblock URL \url{https://www.youtube.com/watch?v=KWmTX9QotGk}.

\bibitem[Schaff et~al.(2019)Schaff, Yunis, Chakrabarti, and Walter]{schaff2019jointly}
Charles Schaff, David Yunis, Ayan Chakrabarti, and Matthew~R Walter.
\newblock Jointly learning to construct and control agents using deep reinforcement learning.
\newblock In \emph{2019 International Conference on Robotics and Automation (ICRA)}, pages 9798--9805. IEEE, 2019.

\bibitem[Schaul et~al.(2011{\natexlab{a}})Schaul, Glasmachers, and Schmidhuber]{schaul2011high}
Tom Schaul, Tobias Glasmachers, and J{\"u}rgen Schmidhuber.
\newblock High dimensions and heavy tails for natural evolution strategies.
\newblock In \emph{Proceedings of the 13th annual conference on Genetic and evolutionary computation}, pages 845--852, 2011{\natexlab{a}}.

\bibitem[Schaul et~al.(2011{\natexlab{b}})Schaul, Glasmachers, and Schmidhuber]{snes}
Tom Schaul, Tobias Glasmachers, and J\"{u}rgen Schmidhuber.
\newblock High dimensions and heavy tails for natural evolution strategies.
\newblock In \emph{Proceedings of the 13th Annual Conference on Genetic and Evolutionary Computation}, GECCO '11, page 845–852, New York, NY, USA, 2011{\natexlab{b}}. Association for Computing Machinery.
\newblock ISBN 9781450305570.
\newblock \doi{10.1145/2001576.2001692}.
\newblock URL \url{https://doi.org/10.1145/2001576.2001692}.

\bibitem[Seita et~al.(2021)Seita, Florence, Tompson, Coumans, Sindhwani, Goldberg, and Zeng]{seita2021learning}
Daniel Seita, Pete Florence, Jonathan Tompson, Erwin Coumans, Vikas Sindhwani, Ken Goldberg, and Andy Zeng.
\newblock Learning to rearrange deformable cables, fabrics, and bags with goal-conditioned transporter networks.
\newblock In \emph{2021 IEEE International Conference on Robotics and Automation (ICRA)}, pages 4568--4575. IEEE, 2021.

\bibitem[Seita et~al.(2023)Seita, Florence, Tompson, Coumans, Sindhwani, Goldberg, and Zeng]{seita2023learning}
Daniel Seita, Pete Florence, Jonathan Tompson, Erwin Coumans, Vikas Sindhwani, Ken Goldberg, and Andy Zeng.
\newblock Learning to rearrange deformable cables, fabrics, and bags with goal-conditioned transporter networks, 2023.

\bibitem[Shi et~al.(2022)Shi, Xu, Huang, Li, and Wu]{shi2022robocraft}
Haochen Shi, Huazhe Xu, Zhiao Huang, Yunzhu Li, and Jiajun Wu.
\newblock Robocraft: Learning to see, simulate, and shape elasto-plastic objects with graph networks.
\newblock \emph{arXiv preprint arXiv:2205.02909}, 2022.

\bibitem[Shi et~al.(2023)Shi, Xu, Clarke, Li, and Wu]{shi2023robocook}
Haochen Shi, Huazhe Xu, Samuel Clarke, Yunzhu Li, and Jiajun Wu.
\newblock Robocook: Long-horizon elasto-plastic object manipulation with diverse tools.
\newblock \emph{arXiv preprint arXiv:2306.14447}, 2023.

\bibitem[Taylor and Rodriguez(2019)]{taylor2019optimal}
Orion Taylor and Alberto Rodriguez.
\newblock Optimal shape and motion planning for dynamic planar manipulation.
\newblock \emph{Autonomous Robots}, 43:\penalty0 327--344, 2019.

\bibitem[Wang et~al.(2023)Wang, Zheng, Ma, Du, Kim, Spielberg, Tenenbaum, Gan, and Rus]{wang2023diffusebot}
Tsun-Hsuan Wang, Juntian Zheng, Pingchuan Ma, Yilun Du, Byungchul Kim, Andrew Spielberg, Joshua Tenenbaum, Chuang Gan, and Daniela Rus.
\newblock Diffusebot: Breeding soft robots with physics-augmented generative diffusion models.
\newblock \emph{arXiv preprint arXiv:2311.17053}, 2023.

\bibitem[Whitman et~al.(2020)Whitman, Bhirangi, Travers, and Choset]{whitman2020modular}
Julian Whitman, Raunaq Bhirangi, Matthew Travers, and Howie Choset.
\newblock Modular robot design synthesis with deep reinforcement learning.
\newblock In \emph{Proceedings of the AAAI Conference on Artificial Intelligence}, volume~34, pages 10418--10425, 2020.

\bibitem[Wu et~al.(2021)Wu, Ze, Dai, Udipi, Paulino, and Zhao]{wu2021stretchable}
Shuai Wu, Qiji Ze, Jize Dai, Nupur Udipi, Glaucio~H Paulino, and Ruike Zhao.
\newblock Stretchable origami robotic arm with omnidirectional bending and twisting.
\newblock \emph{Proceedings of the National Academy of Sciences}, 118\penalty0 (36):\penalty0 e2110023118, 2021.

\bibitem[Xu et~al.(2021)Xu, Spielberg, Zhao, Rus, and Matusik]{xu2021multi}
Jie Xu, Andrew Spielberg, Allan Zhao, Daniela Rus, and Wojciech Matusik.
\newblock Multi-objective graph heuristic search for terrestrial robot design.
\newblock In \emph{2021 IEEE international conference on robotics and automation (ICRA)}, pages 9863--9869. IEEE, 2021.

\bibitem[Xu et~al.(2022)Xu, Chi, Burchfiel, Cousineau, Feng, and Song]{xu2022dextairity}
Zhenjia Xu, Cheng Chi, Benjamin Burchfiel, Eric Cousineau, Siyuan Feng, and Shuran Song.
\newblock Dextairity: Deformable manipulation can be a breeze.
\newblock \emph{arXiv preprint arXiv:2203.01197}, 2022.

\bibitem[Yadav et~al.(2021)Yadav, Cuccia, Virot, Rubinstein, and Gerasimidis]{yadav2021nondestructive}
Kshitij~Kumar Yadav, Nicholas~L Cuccia, Emmanuel Virot, Shmuel~M Rubinstein, and Simos Gerasimidis.
\newblock A nondestructive technique for the evaluation of thin cylindrical shells' axial buckling capacity.
\newblock \emph{Journal of Applied Mechanics}, 88\penalty0 (5):\penalty0 051003, 2021.

\bibitem[Yan and Mehta(2022)]{yan2022cut}
Wenzhong Yan and Ankur Mehta.
\newblock A cut-and-fold self-sustained compliant oscillator for autonomous actuation of origami-inspired robots.
\newblock \emph{Soft Robotics}, 9\penalty0 (5):\penalty0 871--881, 2022.

\bibitem[Yang et~al.(2021)Yang, Vella, and Holmes]{kirigami2021}
Yi~Yang, Katherine Vella, and Douglas~P. Holmes.
\newblock Grasping with kirigami shells.
\newblock \emph{Science Robotics}, 6\penalty0 (54):\penalty0 eabd6426, 2021.
\newblock \doi{10.1126/scirobotics.abd6426}.
\newblock URL \url{https://www.science.org/doi/abs/10.1126/scirobotics.abd6426}.

\bibitem[Ze et~al.(2022)Ze, Wu, Dai, Leanza, Ikeda, Yang, Iaccarino, and Zhao]{ze2022spinning}
Qiji Ze, Shuai Wu, Jize Dai, Sophie Leanza, Gentaro Ikeda, Phillip~C Yang, Gianluca Iaccarino, and Ruike~Renee Zhao.
\newblock Spinning-enabled wireless amphibious origami millirobot.
\newblock \emph{Nature communications}, 13\penalty0 (1):\penalty0 3118, 2022.

\bibitem[Zhao et~al.(2020)Zhao, Xu, Konakovi{\'c}-Lukovi{\'c}, Hughes, Spielberg, Rus, and Matusik]{zhao2020robogrammar}
Allan Zhao, Jie Xu, Mina Konakovi{\'c}-Lukovi{\'c}, Josephine Hughes, Andrew Spielberg, Daniela Rus, and Wojciech Matusik.
\newblock Robogrammar: graph grammar for terrain-optimized robot design.
\newblock \emph{ACM Transactions on Graphics (TOG)}, 39\penalty0 (6):\penalty0 1--16, 2020.

\end{thebibliography}

\end{document}